\documentclass[twocolumn]{article}

\usepackage{fullpage}
\usepackage{graphicx}
\usepackage{hyperref}
\usepackage{caption}
\usepackage{float}
\usepackage{cuted} % for full-width blocks in two-column mode
\usepackage{titling}          % lets us adjust space around \maketitle

\preauthor{\begin{center}\small}
\postauthor{\par\end{center}\vspace{-0.9em}}% <<< key: shrink space after authors/affils ONLY

\predate{}                                   % no date line
\postdate{\vspace{-1.0em}}                   % extra tighten if \date{} still allocates space

\usepackage{titlesec}
\titlespacing*{\section}      {0pt}{*1.4}{0.25\baselineskip}  % before / after
\titlespacing*{\subsection}   {0pt}{*1.1}{0.2\baselineskip}
\titlespacing*{\subsubsection}{0pt}{*1.0}{0.15\baselineskip}

\usepackage{abstract}
\title{From Limited Data to Rare-event Prediction: LLM-powered Feature Engineering and Multi-model Learning in Venture Capital}
\usepackage{authblk}
\date{}   % hides the date

\pretitle{\begin{center}\LARGE}
\posttitle{\par\end{center}\vspace{-0.4em}} % less space after the title

\preauthor{\begin{center}\small}
\postauthor{\par\end{center}\vspace{-0.5em}}% <<< key: shrink space after authors/affils ONLY

\predate{}                                   % no date line
\postdate{\vspace{-1.0em}}                   % extra tighten if \date{} still allocates space

\makeatletter
\renewcommand\AB@affilsepx{, \protect\Affilfont}
\makeatother

\renewcommand\Affilfont{\small}     % keeps affiliations small too

\usepackage{authblk}

\makeatletter
\renewcommand\AB@affilsepx{, \protect\Affilfont}
\makeatother

\renewcommand\Affilfont{\normalsize}    % bigger affiliations

\author[1]{Mihir Kumar}
\author[2]{Aaron Ontoyin Yin}
\author[2]{Zakari Salifu}
\author[2]{Kelvin Amoaba}
\author[2]{Afriyie Kwesi Samuel}
\author[2]{Fuat Alican}
\author[2]{Yigit Ihlamur}

\affil[1]{University of Oxford}
\affil[2]{Vela Research}

\begin{document}

\maketitle
\begin{strip}
\centering
\begin{minipage}{0.96\textwidth}
\begin{abstract}
This paper presents a framework for predicting rare, high-impact outcomes by integrating large language models (LLMs) with a multi-model machine learning (ML) architecture. The approach combines the predictive strength of black-box models with the interpretability required for reliable decision-making. We use LLM-powered feature engineering to extract and synthesize complex signals from unstructured data, which are then processed within a layered ensemble of models including XGBoost, Random Forest, and Linear Regression. The ensemble first produces a continuous estimate of success likelihood, which is then thresholded to produce a binary rare-event prediction. We apply this framework to the domain of Venture Capital (VC), where investors must evaluate startups with limited and noisy early-stage data. The empirical results show strong performance: the model achieves precision between 9.8× and 11.1× the random classifier baseline in three independent test subsets. Feature sensitivity analysis further reveals interpretable success drivers: the startup's category list accounts for 15.6\% of predictive influence, followed by the number of founders, while education level and domain expertise contribute smaller yet consistent effects. 
\end{abstract}
\end{minipage}
\end{strip}

\section{Introduction}
  Predicting rare, high-impact outcomes is a central challenge in machine learning, particularly when data is limited and noisy. In venture capital (VC), this problem is particularly acute: investors must evaluate the likelihood of startup success at the earliest stages, under conditions of high uncertainty. Therefore, precision is critical, and machine learning (ML) models must maximize signal extraction from sparse information. Large language models (LLMs) enhance this capability by capturing patterns in unstructured data that traditional ML methods struggle to reflect.

The objective of this paper is to develop a pipeline that maximizes predictive precision while retaining interpretability. To account for differences in base success rates across datasets, we measure precision relative to a random classifier, targeting performance at least an order of magnitude higher. Our approach combines LLM-powered feature engineering with a multi-model learning architecture. LLMs are used to derive a rich set of features from limited and unstructured data, while the ensemble of models, including XGBoost, Random Forest, and Linear Regression, integrates these features into a continuous prediction of total funding. This funding estimate is then used to predict binary success outcomes. The pipeline is designed to achieve precision over 10× higher than the random baseline while maintaining recall above 10\%, ensuring that improvements in accuracy are not achieved at the expense of coverage.

  \section{Related Work}
Recent advances in LLMs have introduced novel methods for modeling in VC and other decision-making domains, particularly through their ability to extract as well as organize complex information. Despite this progress, balancing predictive accuracy with interpretability continues to be a key challenge.

GPTree \cite{gptree} combines the structured clarity of decision trees with LLM-driven reasoning, achieving a precision of 9.4×. Our approach extends this work by employing a broader multi-model architecture and leveraging LLMs for feature engineering to enrich the dataset and further improve predictive precision.

Founder-GPT \cite{map} proposes a framework that evaluates the fit between founders and their ideas through iterative LLM-based reasoning, and uses these evaluations to improve the predictions of startup success. We adopt a related perspective by employing LLMs for feature engineering in our dataset, allowing founder–idea characteristics to be systematically represented as predictive features.

Random Rule Forest (RRF) \cite{rrf} introduces an ensemble framework in which LLM-generated yes/no questions serve as heuristics. By filtering, grouping, and weighting these heuristics, RRF constructs a lightweight and transparent decision-making model that achieves 5.4× precision.

Our framework builds on these previous approaches, but adopts a different strategy to leverage the capabilities of LLMs. Rather than using LLMs solely to generate predictions or decision rules, we employ them to perform feature selection and engineering at scale. This enables the construction of a richer and more predictive feature space, incorporating not only structured attributes but also qualitative dimensions of founders and startups that are otherwise difficult to encode.
\section{Methods}
Our methodology combines three components: (i) a curated dataset of startup founders, (ii) LLM-powered feature engineering to convert unstructured founder data into structured predictors, and (iii) a multi-model learning architecture that integrates these features to forecast both funding levels and ultimate success. 
\subsection{Dataset}
The dataset consists of 10,825 startup founders and detailed information about their ventures. Each founder profile includes education, professional experience, prior startup experience, and other publicly available attributes, complemented by additional information extracted using LLM-based enrichment techniques. Companies that raised between \$100K and \$4M are categorized as unsuccessful. Founders are labeled as successful if their company achieved a valuation above \$500M through an initial public offering (IPO), was acquired for more than \$500M, or raised over \$500M in funding. Under these criteria, 922 founders (8.5\%) fall into the successful class, while the remaining 9,903 are classified as unsuccessful.
\subsection{Feature Engineering Methods}
The creation of sophisticated features for training required an extensive engineering process largely driven by LLMs, following the approach of Ozince and Ihlamur (2024) \cite{autoVC}. This enabled the construction of a far broader feature set than would be feasible with traditional methods. For example, we extracted a Skill Relevance variable, encoded on a scale from 0 (no relevance) to 4 (high relevance), which captures the alignment between a founder’s skills and the startup’s domain. Features of this kind are difficult to generate without LLMs, as they require natural language processing and reasoning beyond conventional pipelines. In total, the LLM-derived process produced 63 trainable features, which were organized into categorical, textual, continuous, and boolean groups for preprocessing.
\subsubsection{Categorical Data}
After parsing the data, several categorical features of the founder were encoded in the model by mapping to an integer following Xiong and Ihlamur (2024) \cite{map}. For example, each founder's education level was mapped through the scheme in Table 1.

\begin{table}[h!tbp]
     \centering
     \begin{tabular}{||c|c||}
          
            \hline
            Highest Education Level & Integer Map  \\ [0.5ex] 
            \hline\hline
            Associate Degree or less & 0 \\ 
            \hline
            Bachelor's Degree & 1\\
             \hline
            Master's Degree & 2\\
            \hline
            Doctoral Degree or more & 3\\
            \hline
     \end{tabular}
     \caption{\textit{Integer mapping of founder education level }(\textit{education\_level})}
     \label{tab:my_label}
 \end{table}
 Furthermore, due to the use of the LLM, further insights into a founder's profile are engineered into categorical data that are used by the model. For example, the \textit{domain\_expertise} variable shown in Table 2 is a representation of how well the founder's industry experience and education match the startup's domain. While this is difficult for traditional ML models to encode as a variable, the LLM allows us to accurately encode this feature into one of 4 categories.
    
    \begin{table}[h!tbp]
        \centering
        \begin{tabular}{||c|c||}
            \hline
            Domain Expertise & Integer category  \\ [0.5ex] 
            \hline\hline
            No alignment & 0 \\ 
            \hline
            Weak Alignment & 1\\
            \hline
             Moderate Alignment & 2\\
            \hline
            Strong Alignment & 3\\
            \hline
        \end{tabular}
        \caption{\textit{LLM encoding of} \textit{domain\_expertise} \textit{variable from the unstructured founder data}}
        \label{tab:my_label}
    \end{table}
\subsubsection{Textual Data}
Startup descriptions were converted into embeddings for use in the model. These embeddings provided a compact representation of unstructured text that could be integrated alongside other feature types.
\subsubsection{Continuous and Boolean Data}
Boolean features do not require additional encoding since their values are already represented as 0 and 1. Continuous features are standardized using Z-scores, which improves the performance of gradient-based algorithms \cite{grad} such as XGBoost and facilitates a more reliable analysis of feature sensitivity.

\subsection{Multi-Model Construction}
\begin{figure}[b!tbh]
    \centering
    \includegraphics[width=0.75\linewidth]{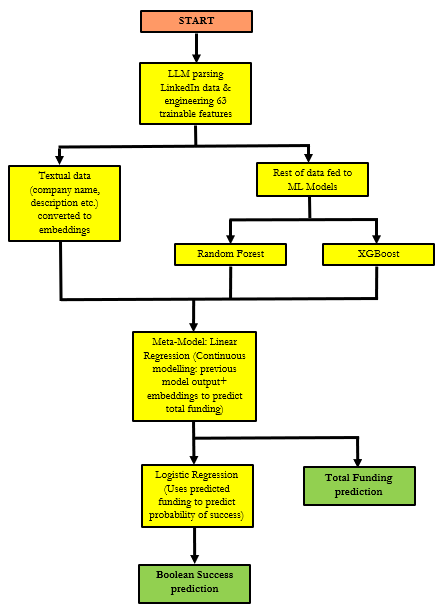}
    \caption{\textit{Multi-model architecture}}
    \label{fig:multi-model}
\end{figure}

After converting each data branch into processable features, we applied a multi-model architecture to predict startup success. The structure of this architecture is illustrated in Figure~\ref{fig:multi-model}.

The first layer consists of two machine learning models and an LLM used to process the data. This layer was designed to maximize precision by combining models that are well suited to different data characteristics. XGBoost and Random Forest were selected for their complementary strengths: XGBoost excels at handling large and complex datasets, while Random Forest is more robust to overfitting after preprocessing \cite{RFXG}. In addition, Random Forest provides stronger interpretability, which complements the advanced predictive capacity of XGBoost.

The outputs of these models, together with the text embeddings, are passed to a meta-model that is based on Linear Regression. This step refines the prediction by introducing additional parameters, and produces a continuous estimate of the total funding a startup is likely to raise. The funding prediction then serves two purposes. First, it improves precision when converted into a binary outcome through logistic regression. Second, it provides insight into feature sensitivity. By analyzing the continuous outputs, startups can be segmented into funding classes with associated probabilities of success, while the relative weighting of features offers interpretable signals for decision-making.

The predicted funding outcomes are mapped to binary success through logistic regression, which encodes the funding prediction into a probability of boolean success. The classification threshold is tuned to maximize precision while avoiding overfitting \cite{threshold}. Because successful startups are rare and investment stakes are large, the threshold was set higher than the conventional 0.5. Model performance was then evaluated in terms of recall and, most importantly, precision, expressed as a multiple of the success rate of a random classifier, represented as \textit{Baseline Rate} throughout the paper. 
\section{Results and Evaluation}
The model was trained on 8,659 instances and evaluated on a held-out set of 2,166 instances. The evaluation set was partitioned into three disjoint test subsets of 722 organizations each, with the number of successes and the corresponding baseline success rates summarized in Table 3.

\begin{table}[h!tbp]
    \centering
    
\begin{tabular}{||c |c|c|c||} 
 \hline
 Subset & Size & Successes & Baseline Rate \\ [0.5ex] 
 \hline\hline
 1 & 722& 57 & 7.9\% \\ 
 \hline
 2 & 722 & 49 & 6.8\%\\
 \hline
 3 & 722 & 67 & 9.3\%\\
 \hline

\end{tabular}
\caption{\textit{The detailed view of subsets}}
    \label{tab:my_label}
\end{table}

\subsection{Funding Prediction}
Funding prediction performance was evaluated using the mean absolute percentage error (MAPE), which quantifies the average relative deviation between predicted and actual funding values. MAPE was chosen because it provides a scale-independent measure of error that is interpretable across organizations of different funding sizes. The results for each subset, as well as for the overall set, are reported in Table 4.

\begin{table}[h!tbp]
    \centering
    \begin{tabular}{||c|c|c||}
        \hline
        Subset & MAPE \\ [0.5ex] 
        \hline\hline
        1 & 3.32\% \\ 
        \hline
        2 & 3.02\% \\
        \hline
        3 & 3.89\%  \\
        \hline
    \end{tabular}
    \caption{\textit{Mean Absolute Percentage Error for funding prediction}} 
    \label{tab:performance}
\end{table}
 
Across all test sets, the MAPE remains below 4\%, indicating that the model predicts funding with high accuracy even under the noisy conditions of early-stage startups.

\subsection{Success Prediction}
To evaluate success prediction, we focused on precision, defined as the proportion of true positives among predicted positives. The logistic regression threshold was varied to examine its effect on precision, following the approach of Handoyo et al. (2021).\footnote{Method adapted from Handoyo, S., Chen, Y. et al., 2021} Rather than fine-tuning, thresholds were tested at fixed intervals to reduce the risk of overfitting. The resulting precision curve is presented in Figure 2.

\begin{figure}[h!tbp]
    \centering
    \includegraphics[width=0.65\linewidth]{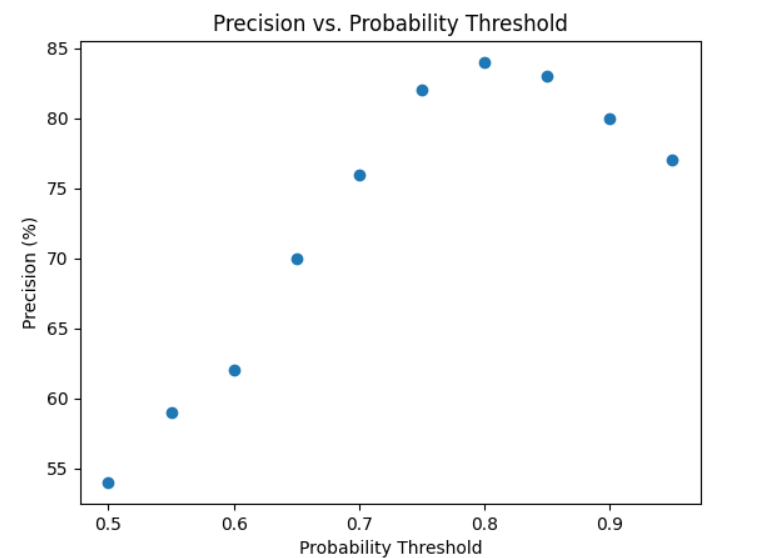}
    \caption{ \textit{Precision (\%) versus threshold probability}}
    \label{fig:enter-label}
\end{figure}

As shown in Figure 2, precision peaks when the classification threshold lies between 0.75 and 0.85. Because model performance within this interval is stable, we selected 0.8 as a representative threshold to avoid overfitting to a specific dataset. Using this threshold, the multi-model was evaluated on each test subset, with precision and recall reported in Table 5.
In each test subset, the multi-model achieved precision more than nine times higher than the baseline success rate with sample 2 achieving over eleven times the base success rate, underscoring its strong predictive performance, while also maintaining recall above 30\%. 

\begin{table}[h!tbp]
    \centering
    \begin{tabular}{||c|c|c||}
        \hline
        Subset & Recall & Precision \\ [0.5ex] 
        \hline\hline
        1 & 36\% & 10.4X \\ 
        \hline
        2 & 35\% & 11.1X \\
        \hline
        3 & 38\% & 9.8X  \\
        \hline
        Overall & 36\% & 10.3X \\
        \hline
    \end{tabular}
    \caption{\textit{Performance of multi-model against the random baseline rate}} 
    \label{tab:performance}
\end{table}

Building on these results, we further segmented startups into funding classes, using the continuous funding prediction to assign probabilities of success for each class of startups. The outcomes of this analysis are reported in Table 6.

\begin{table}[h!tbp]
    \centering
    \begin{tabular}{||c|c||}
        \hline
        Funding Class(\$) & Probability of Success (\%)\\
        \hline\hline
         100K-1M & 1.27\%\\
         \hline
         1M-10M & 8.41\%\\
         \hline
         10M- 100M & 80.89\%\\
         \hline
         100M-1B & 95.35\%\\
         \hline
         1B+ & 100.00\%\\
         \hline
    \end{tabular}
    \caption*{Table 6: \textit{Probabilities of success for each funding class}\footnote{Funding classes adapted from Vela Partners’ GitHub}}
    \label{tab:my_label}
\end{table}

\subsection{Feature Sensitivity Analysis}
Feature sensitivity was evaluated using the parameter weights from each component model in the multi-model architecture, combined with the contributions of the text embeddings. Each feature’s influence was quantified as a percentage of total predictive weight. To ensure robustness, the model was re-tested on multiple subsets with varying proportions of outliers, which yielded a consistent hierarchy of feature importance. The relative contributions of the ten most impactful features are shown in Figure 3.  

Among these, the category list (the set of fields in which the startup operates) and the number of founders emerge as the strongest predictors. By contrast, education level, while still appearing in the top ten, accounts for only about one tenth of the weight assigned to the number of founders.

\begin{figure}[b!htp]
    \centering
    \includegraphics[width=0.65\linewidth]{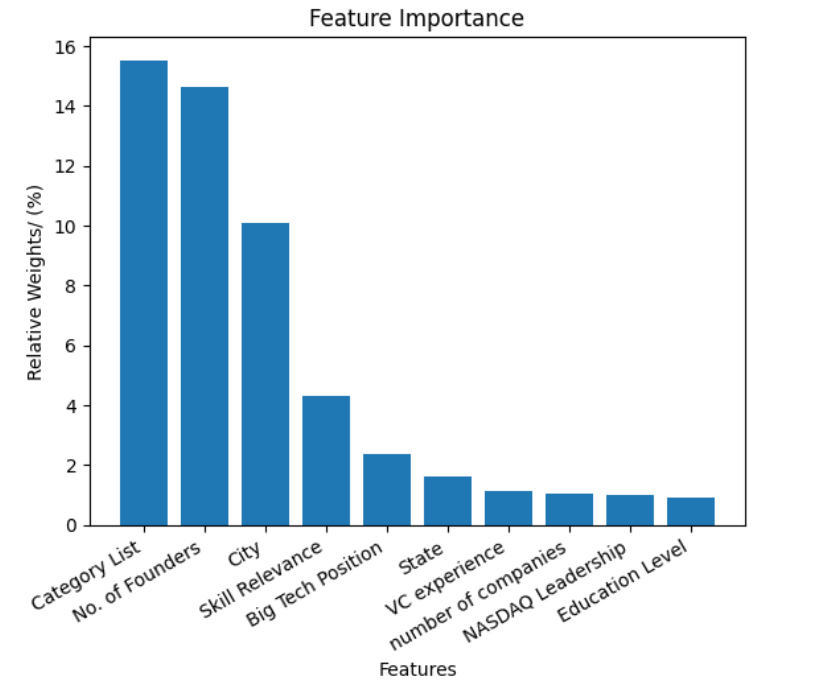}
    \caption{\textit{Bar chart comparing the relative importance of each feature in predicting a startup's success} }
    \label{fig:enter-label}
\end{figure}

 \subsection{Ablation Studies}
To assess the contribution of individual components and design choices, we performed a series of ablation experiments. These examined the impact of LLM-powered features and feature categories, embedding sources, and the ensemble structure. 

\subsubsection{LLM-based vs. Traditional Features}
We removed the LLM-engineered features, leaving only deterministic variables derived from standard structured information (38 features in total). As reported in Table 7, running the same pipeline on an equivalent train/test split resulted in a precision drop from 10.4× to 4.6×. This substantial decrease highlights the importance of semantic information captured by the LLM, particularly in encoding founder-specific attributes that are not well represented by traditional features.
\begin{table}[h!tbp]
    \centering
    \begin{tabular}{||c|c|c|c||}
        \hline
        Sample size & Baseline rate & Precision & Recall\\
        \hline\hline
         722 & 7.9\% & 4.6X & 26\%\\
         \hline
    \end{tabular}
    \caption*{Table 7: \textit{Performance without LLM features}}
    \label{tab:my_label}
\end{table}

\subsubsection{Embedding Comparison}
We compared model performance using text embeddings from \texttt{text-embedding-ada-002}, \texttt{all-MiniLM-L6-v2}, and a baseline without embeddings. As seen in Table 8, LLM-based embeddings achieved the highest precision, while MiniLM remained competitive, and both outperformed the no-embedding baseline.
\begin{table}[h!tbp]
    \centering
    \begin{tabular}{||c|c|c||}
        \hline
        Embedding & Precision & Recall\\
        \hline\hline
         ada-002 & 10.4X & 36\%\\
         \hline
          MiniLM& 10.2X & 36\%\\
         \hline
          None & 8.7X & 30\%\\
         \hline
    \end{tabular}
    \caption*{Table 8: \textit{Embedding model comparisons}}
    \label{tab:my_label}
\end{table}

\subsubsection{Model Architecture Variants}
We systematically evaluated the pipeline by removing each component of the multi-model architecture in turn. Table 9 showed that removing XGBoost from the pipeline produced the largest drop in performance, underscoring its importance for handling high-dimensional, multi-category data in this setting. Replacing the Linear Regression meta-model with a shallow Neural Network yielded higher precision and recall but introduced overfitting, leading to inconsistent and unreliable outcomes. In addition, feature sensitivity analysis became less stable when either the Random Forest or Linear Regression components were removed, reducing the consistency of the ranking of high-impact features.
\begin{table}[b!thp]
    \centering
    \begin{tabular}{||c|c|c||}
        \hline
        Model&Precision & Recall \\
        \hline\hline
         XGBoost & -3.2X & -7\%\\
         \hline
          Random Forest& -2.3X & -5\%\\
         \hline
          Meta-model & +0.4X & +2\%\\
         \hline
    \end{tabular}
    \caption*{Table 9: \textit{The impact of components on model performance}}
    \label{tab:my_label}
\end{table}

\subsubsection{Feature Category Removal}
We further evaluated the contribution of feature engineering by retraining the model on the same train–test split while systematically removing each data subset (textual, categorical, continuous, and boolean) from training.

The ablation results in Table 10 show that categorical data produced the largest decline in model performance. Since many of the categorical features were generated through LLM-powered engineering, this result emphasizes the role of LLM-derived features in driving both precision and recall.
\begin{table}[h!tbp]
    \centering
    \begin{tabular}{||c|c|c||}
        \hline
        Feature Type  & Precision & Recall \\
        \hline\hline
         Categorical & -3.9X & -9\%\\
         \hline
          Continuous & -3.0X & -6\%\\
         \hline
          Boolean & -1.7X & -5\%\\
         \hline
         Textual & -1.7X & -6\%\\
         \hline
    \end{tabular}
    \caption*{Table 10: \textit{The impact of feature types on model performance}}
    \label{tab:my_label}
\end{table}

\section{Conclusion}
We presented a multi-model framework that integrates ML methods with LLM-powered feature engineering to predict startup success. The pipeline was evaluated on funding prediction, binary success classification, and feature sensitivity analysis, consistently demonstrating high precision and robustness. In particular, the model achieved over 10× the baseline precision while maintaining interpretable feature attributions, showing that it can deliver both predictive power and transparency. These results highlight the potential of combining LLM-derived features with ensemble learning to address rare-event prediction tasks such as early-stage startup investing evaluation, where traditional models often fail to generalize.

\subsection{Limitations}
Although our approach achieves high precision and yields interpretable feature sensitivities, several limitations remain. 

The layered design of the continuous funding predictor followed by logistic regression introduces error propagation: inaccuracies in the funding estimate, though small (below 4\% MAPE), may still influence the calibrated probabilities used for success classification. 

In addition, reliance on LLMs for feature engineering introduces risks, as all 63 features are LLM-derived and subject to misclassification, particularly for variables with subjective boundaries such as skill relevance. 

Finally, the model’s performance is constrained by the quality and coverage of the underlying data. Founder-centric profiles constructed from publicly available sources with information provided by founders themselves may embed biases toward individuals with greater online presence, limiting the generalizability of results across different ecosystems.

\subsection{Future Directions}
Future work can be done to address some of these limitations and to expand the scope of this research. 

Refining the multi-model architecture and exploring model combinations could reduce error propagation. 

Additional research into feature engineering is also needed, particularly methods that can validate or augment LLM-derived features. In this regard, integrating SHAP-based explanations with LLM outputs could provide a systematic way to quantify feature contributions while exposing potential inconsistencies. Likewise, developing methods to detect, measure, and reduce hallucinations in LLM-driven feature generation will be important to improve reliability.

\end{document}